\documentclass[11pt]{article}
\usepackage{acl}

\usepackage{fontspec}
\usepackage{polyglossia}
\setdefaultlanguage{english}
\setotherlanguage{belarusian}

\usepackage{latexsym}
\usepackage{microtype}
\usepackage{graphicx}
\usepackage{booktabs}
\usepackage{float}
\usepackage{multirow}
\usepackage{placeins}
\usepackage{enumitem}
\usepackage{array}
\title{ReMova: Fine-tuning LLMs for English to Belarusian translation\thanks{The name combines the prefix \emph{re-} with
  Belarusian \texttt{мова} (`language').}}
\renewcommand{\arraystretch}{1.2}

\author{Mikita Pilinka$^1$
\And
Aliaksandr Kliujeŭ$^2$
\And
David Samuel$^1$
\And
Yves Scherrer$^1$
\AND
\vspace{-3em} \phantom{x} \\ $^1$ University of Oslo \\  Department of Informatics \\
\texttt{\{mikitap,davisamu,yvessc\}@uio.no}
\And
\vspace{-3em} \phantom{x} \\ $^2$ Independent Researcher \\ orcid.org/0009-0005-0720-8175 \\ \texttt{katoshrodingera@protonmail.ch}
}

\begin{document}
\maketitle
\begin{abstract}
This paper presents a Belarusian-specific data-cleaning pipeline and fine-tuning for English-Belarusian machine translation. 
Our cleaning pipeline distinguishes itself from others by employing a 
correction tool that addresses the issue of the two orthographies of the Belarusian language, noise in the training data, interference from other languages and other misspelling issues common in Belarusian on the internet. 
A matched ablation on unfiltered training data shows substantial benefits from filtering for all fine-tuned models, with the LLM-based models gaining roughly twice as much from filtering as the dedicated encoder-decoder MT system, supporting the view that for Belarusian MT one of the primary bottlenecks is data quality.
\end{abstract}

\section{Introduction}

The uneven distribution of languages across the web and the wide variation in available linguistic resources have left many languages, including Belarusian, under-prioritised in the training and evaluation of machine translation systems. We present a Belarusian-specific data-cleaning pipeline and use the resulting corpus to fine-tune translation models from two neural MT paradigms: NLLB~\cite{nllbteam2022languageleftbehindscaling} as an encoder-decoder system, Gemma-4~\cite{gemmateam2026gemma4technicalreport} and TranslateGemma~\cite{finkelstein2026translategemmatechnicalreport} as decoder-only LLMs.
We release our primary model, the fine-tuned TranslateGemma-12B-IT, under the Gemma Terms of
Use.\footnote{\url{https://huggingface.co/mineralsfree/ReMova-TranslateGemma-en-be}}
\subsection{Belarusian variation}

Contemporary written Belarusian displays genuine orthographic variation, having two active graphical codified systems: cyrillic (mostly) and latin (rarely), 
and featuring two competitive normative systems: the official standard orthography (often informally called
\textit{Narkamaŭka}) and the alternative orthography, \textit{Taraškievica}, characterized by lesser codification and the lack of common standardizing authority. Although these names originally referred to different orthographic systems, they subsequently came to denote broader normative systems, thus their
differences extend beyond spelling -- concerning phonetics, morphology, word formation, vocabulary and syntax. The opposition is not entirely categorical, however, because many grammatical forms,  allegedly distinctive for \textit{Taraškievica}, are also admitted, stylistically marked, or used as variants within the official standard~\cite{ramza2018taraskievica}. \citet[p.~513]{aparovich-etal-2025-belarusianglue} found that most digital examples in their sentiment dataset followed the official standard, fewer than 10\% used \textit{Taraškievica}, and a small minority used the Latin alphabet \textit{lacinka}. This work focuses exclusively on standard Belarusian orthography. Extending the model to other Belarusian varieties will require dedicated evaluation and training data to ensure adequate, variety-specific performance.

\paragraph{Language interference.}
The existence of several orthographic traditions in Belarusian complicates the formation of coherent training data. As a result, a model may learn from heterogeneous linguistic material, creating noise in its outputs. As \citet[pp.~18--19, 24--27]{nazaranka2025Naviga} notes, this may manifest itself, for example, in the coexistence of elements from two orthographies within a single text, the use of erroneous constructions at different levels of language, or unstable switching between different norms. An additional problem is interference from other closely related languages, particularly Russian and, according to our observations, Ukrainian and Polish. These languages may fill gaps resulting from the limited amount of Belarusian-language training data. Such interference may occur not only at the lexical level but also at the phonetic, morphological, syntactic, and stylistic levels.

It is worth noting that a significant portion of the modern Belarusophone community consists of ``new speakers'' raised in Russian-speaking families  \cite[pp.~63--65]{woolhiser2013new}. In Woolhiser's pilot sample, performed in 2013, most of these speakers reported shifting to Belarusian in peer-group communication during adolescence or early adulthood, between the ages of 13 and 22 \cite[pp.~86--87]{woolhiser2013new}. Their usage coexists with the two above-mentioned normative systems and displays variation associated with speakers' differing preferences for forms perceived as more or less divergent from Russian \cite[pp.~107--111]{woolhiser2013new}. Considering current trends in Belarusian language, we suspect that the proportion of ``new speakers'', that learned Belarusian as the second language, has only grown since the date of Woolhiser's work.

\paragraph{Misspellings.}

Orthographic diversity and language interference should be distinguished from misspellings, pseudo-Belarusian forms, and unreviewed machine translation.

\citet{quovadis2025} identify the low level of proficiency for literary Belarusian, together with the desire to produce texts with minimal effort and an insufficient understanding of automatic translation, as factors contributing to the growing presence of machine-translated language in texts of different genres. According to their analysis, artificial and unnatural forms generated by the early Russian-to-Belarusian translation system \textit{Belazar} entered media texts and, through repetition, gradually began to be perceived as correct. The subsequent use of \textit{Google Translate} and \textit{Yandex Translate} intensified this process, with traces of automatic translation reportedly increasing rapidly across genres. The effects described are not limited to isolated spelling errors. Corpus research reports that machine-translated texts introduce artificially formed lexical items and grammatical constructions into ordinary usage and may produce semantic shifts. This calls into question the value of some Belarusian-language textual data and the extent to which such language can be considered representative for research purposes \cite[p.~146--147]{quovadis2025} and, by extension, for creating a proper Belarusian dataset.

\citet[p.~56]{liasovich2024korpusnajalingvistyka} offers a different view, arguing that media texts published online reflect the current state of the Belarusian language’s lexical resources and their ongoing transformation, thereby revealing the condition of contemporary living Belarusian.

\subsection{Importance of data cleaning}
Looking at the available parallel data, we also noticed the high degree of variability of modern Belarusian language, which emphasizes the need of a proper data cleaning process that distinguishes normative systems, and cleans obvious errors and language interference issues. We propose a balanced approach that does not step into the internal processes of the language change -- not imposing subjective views on how the language should change in the future. In our work we try to stabilise the lexical, orthographical and typographical features of the text, according to the standard Belarusian orthography, while accepting new word-forms, without interfering at the syntactical level of the written texts, although we want to mark that syntax should be the matter of future works in this field.
\begin{figure*}[t]
  \centering
  \includegraphics[width=\linewidth]{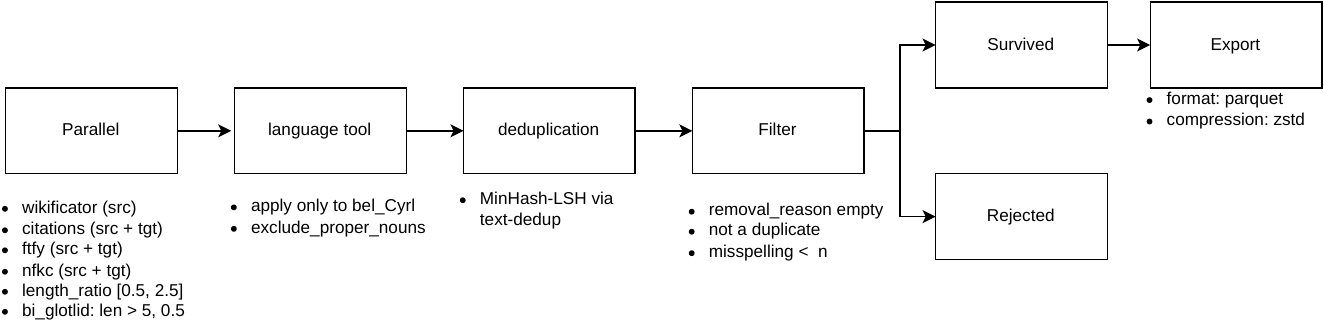}
  \caption{The multiple stages of the \textsc{ReMova} cleaning pipeline employed to ensure the quality of the English-Belarusian parallel dataset.}
  \label{fig:pipeline}
\end{figure*}

\section{Methodology}
\label{sec:methodology}

In this study, we use a custom filtering pipeline, which we refer to as \textsc{ReMova},\footnote{ReMova github: \url{https://github.com/mineralsfree/ReMova}} that combines current tooling and quality metrics to extract reliable parallel segments. 
The pipeline is organised into stages, each of which writes its output to a file and records summary statistics. 
Some stages apply to both sides of the parallel data, such as ftfy,\footnote{ftfy (\emph{fixes text for you}): \url{https://github.com/rspeer/python-ftfy}.} NFKC normalisation,\footnote{NFKC: Unicode Normalisation Form Compatibility Composition; see \url{https://unicode.org/reports/tr15/}.} and GlotLID-based language identification,\footnote{GlotLID: \url{https://github.com/cisnlp/GlotLID}.} while others address the aforementioned issues specific to the Belarusian side (Wikificator\footnote{Wikificator: \url{https://be.wikipedia.org/wiki/Вікіпедыя:Вікіфікатар}.} and LanguageTool\footnote{LanguageTool: \url{https://languagetool.org/}.}).
A schematic representation of the pipeline is shown in Figure~\ref{fig:pipeline}.

\subsection{Belarusian-specific normalisation}

A central element of our approach is the strict handling of misspellings and \textit{Taraškievica} forms. We rely on Wikificator~\cite{wiki_wikificator_be}, a set of regular expressions maintained by the Belarusian Wikipedia community to detect and correct the most common orthographic errors and \textit{Taraškievica} variants found in Belarusian text. We complement this with LanguageTool~\cite{languagetool}, an open-source proofreading platform whose Belarusian module provides a spellchecker based on the grammatical base of the National Corpus of the Belarusian Language~\cite{koshchanka2018bnkorpus}. The National Corpus of the Belarusian Language, which is compiled by the Institute of Belarusian Language, currently recommends LanguageTool as the main spellchecker solution for the Belarusian language and contributes to the project~\cite{bnkorpus_spellchecker}. We therefore considered it the most advanced and reliable solution for our spell-checking workflow. While Wikificator targets a fixed set of high-frequency patterns, LanguageTool offers broader dictionary-based coverage, catching misspellings and grammatical inconsistencies that fall outside the regex rules.

The different stages of the pipeline have distinct roles. Most either normalise text in place (ftfy, NFKC, Wikificator) or annotate each segment with metrics that downstream stages consume: GlotLID adds language-identification confidence, LanguageTool tags the Belarusian side with match counts and match density (matches per non-proper-noun word), and deduplication marks near-duplicates. Removal is concentrated at two places. A dedicated filter stage applies thresholds on the accumulated metrics. Before running the pipeline, we discard trivially short pairs (two words or fewer).

\begin{table}[!th]
  \centering
  \small
  \setlength{\tabcolsep}{4pt}
  \begin{tabular}{@{}lll>{\raggedright\arraybackslash}p{3.8cm}@{}}
  \toprule
  \textbf{Stage}        & \textbf{In}     & \textbf{Out}    & \textbf{Notable effect} \\
  \midrule
  prep                  & 8{,}980 & 7{,}899 & 1{,}081 pairs with $\le 2$ words dropped \\
  parallel              & 7{,}899 & 7{,}899 & 271 typography, 201 mojibake, 101 NFKC \\
  languagetool          & 7{,}899 & 7{,}899 & 1{,}051 rows with LT matches (13.3\%) \\
  dedup                 & 7{,}899 & 7{,}899 & near-duplicates marked (signal-only) \\
  filter                & 7{,}899 & 6{,}953 & 88\% survived; 692 LT errors, 231 dup, 23 lang/len \\
  BLASER 
  & 6{,}953 & 6{,}749 & 204 low-quality pairs dropped (2.9\%) \\
  export                & 6{,}749 & 6{,}749 & $\sim$285\,KB parquet (ZSTD) \\
  \bottomrule
  \end{tabular}
  \caption{Tatoeba (Belarusian--English) pipeline execution. 8{,}980 raw pairs $\to$ 6{,}749 training pairs (75\% overall yield).}
  \label{tab:clean-tatoeba}
\end{table}

Table~\ref{tab:clean-tatoeba} shows the effects of the different stages on one of the utilised corpora, Tatoeba. It can be seen that the filter stage dominates the rest: about 12\% of the pairs are dropped there, mostly because they exceed the manually selected LanguageTool density threshold (692 pairs, a proxy for \textit{Taraškievica} influence and chronic misspelling) or were flagged as duplicates (231 pairs). Because the density metric is sensitive to LanguageTool's coverage of Belarusian, we iteratively curated its dictionary: the most frequent matches were either added with their correct spelling or marked as ignorable where the match was a false positive. 

\subsection{Training data}
We selected the training dataset through both human review and
pipeline-based evaluation. Rather than relying on a single corpus,
we combined five sources with per-corpus BLASER~\cite{seamlessm4t2023blaser} cutoffs calibrated
to their register:

\begin{itemize}[noitemsep]
\item HPLT web-crawled bitext \citep{burchell-etal-2025-expanded},
\item OpenSubtitles-derived
paragraph alignments \citep[upsampled 2$\times$ to enrich the spoken
register;][]{lison-tiedemann-2016-opensubtitles2016},
\item Tatoeba crowdsourced pairs \citep[upsampled 2$\times$;][]{tiedemann-2020-tatoeba},
\item Wikimedia aligned paragraphs \citep{tiedemann-2012-parallel}, and
\item back-translated segments
from a curated subset of HPLT's monolingual Belarusian, itself derived from Common Crawl and the Internet Archive (upsampled 2$\times$).
\end{itemize}

\noindent
Table~\ref{tab:training-corpora} summarises the composition.
Per-corpus thresholds were necessary because BLASER's
written-text calibration systematically under-scores subtitle-style
compression even for faithful translations; a uniform cutoff would have
discarded most of the spoken-register signal. Other OPUS subcorpora such
as WikiMatrix were excluded, as many of their text segments
are misaligned.

To strengthen the coverage of informal Belarusian, we augmented the training
set with back-translated pairs from selected Common Crawl web resources. To improve the ability of our models to work with spoken registers we created a whitelist-curated subset of the HPLT monolingual Belarusian collection
(2{,}300 segments after cleaning) in two particular registers: ID (Interactive Discussion) and SP (Spoken). The resulting segments were back-translated into English twice with a large language model DeepSeek-V4-Flash~\cite{deepseekai2026deepseekv4}, yielding 4{,}600 BE-EN pairs whose two
targets are textually distinct in 84\% of cases and thus function as
paraphrastic augmentation.



\section{Experiments}

\subsection{Identifying strongest baseline}

To get a fuller picture, we fine-tune models from both paradigms: an encoder-decoder system and a decoder-only LLM. We target the constrained track which requires releasing a model under 20B parameters under a licence permitting unrestricted non-commercial use~\cite{wmt26_general_mt}. We compare a subset of the suggested base models (Gemma-3~\cite{gemmateam2025gemma3technicalreport}, Aya~\cite{üstün2024ayamodelinstructionfinetuned}, Qwen-2.5~\cite{qwen2025qwen25}) along with Llama-3.2~\cite{grattafiori2024llama3} before fine-tuning to identify the strongest baseline for Belarusian translation. As the encoder-decoder counterparts, we use both NLLB and MADLAD-400~\cite{kudugunta2023madlad400multilingualdocumentlevellarge}, a comparably positioned Google model covering 400 languages.

An initial experiment compared base models under 4B parameters, a cap chosen to allow rapid iteration within the available compute budget. 
Note that no fine-tuning was applied at this stage; the goal was to identify the strongest base model under the parameter budget before committing compute to adapter training.
\subsection{Fine-tuning}
The strongest models were chosen for fine-tuning and further evaluation on the same metrics and dataset: full fine-tuning for the encoder-decoder (NLLB) via \texttt{Seq2SeqTrainer}, and LoRA-based supervised fine-tuning of the LLMs (Gemma-4-12B-IT and TranslateGemma-12B-IT) via TRL's \texttt{SFTTrainer}. Both regimes use AdamW with warmup ratio 0.05; weight decay is 0.01 for NLLB and 0.0 for the LLMs. LoRA is applied to \texttt{all-linear} modules for Gemma-4; TranslateGemma restricts the target set to \texttt{q,k,v,o,gate,up,down} projections to avoid adapting the vision encoder. The parameters used for fine-tuning selected models are listed in Table~\ref{tab:hparams}. Our training corpus mixes sentence pairs (Tatoeba) with paragraph-level and multi-sentence alignments (Wikimedia, OpenSubtitles, HPLT); the chosen \texttt{max\_length} budgets (128 per side for NLLB, 256 for Gemma-4 and TranslateGemma) cover the vast majority of pairs, and
BOUQuET is evaluated at its native sentence granularity.

\begin{table}[!t]
  \centering
  \small
  \setlength{\tabcolsep}{3pt}
  \begin{tabular}{@{}lrrrr@{}}
  \toprule
  \textbf{Corpus}         & B\textbf{LASER $\geq$} & \textbf{Rows} & \textbf{$\times$} & \textbf{Share} \\
  \midrule
  HPLT (web)            & 3.5 & 83{,}770 & 1 & 49.4\% \\
  OpenSubtitles   & 3.0 & 29{,}849 & 2 & 35.2\% \\
  Tatoeba         & 3.5 &  6{,}749 & 2 &  8.0\% \\
  Wikimedia       & 3.7 &  7{,}923 & 1 &  4.7\% \\
  ID/SP (HPLT)           &  -- &  2{,}300 & 2 &  2.7\% \\
  \midrule
  \multicolumn{4}{@{}l}{Total} & \textbf{169{,}489} \\
  \bottomrule
  \end{tabular}
  \caption{Composition of the fine-tuning corpus. Per-corpus BLASER thresholds are calibrated to each source's register; the $\times$ column reports the upsampling factor and \textit{Share} is the resulting share of the combined training set. ID/SP (HPLT) denotes back-translated in-domain parallel text pairs generated from a whitelist-curated subset of HPLT monolingual Belarusian segments.}
  \label{tab:training-corpora}
\end{table}

\begin{table}[!t]
\small
  \centering
  \setlength{\tabcolsep}{4pt}

  \begin{tabular}{@{}lcc@{}}
  \toprule
                                & \textbf{NLLB FT}              & \textbf{LLM LoRA SFT$^\dagger$} \\
  \midrule
  Base model                    & \texttt{nllb-200-3.3B}        & \texttt{gemma-4-12b-it} \\
  Epochs                        & 3                    & 2 \\
  Eff.\ batch size              & 32                   & 16 \\
  Learning rate                 & $3 \times 10^{-5}$   & $1 \times 10^{-4}$ \\
  Max seq length                & 128                  & 256 \\
  LoRA ($r$ / $\alpha$ / drop)        & ---                  & 16 / 32 / 0.05 \\
  \bottomrule
  \end{tabular}
  \caption{Hyperparameters for fine-tuning. $^\dagger$Both Gemma-4-12B-IT and TranslateGemma-12B-IT share this LoRA recipe.}
  \label{tab:hparams}
\end{table}

\subsection{Filtering ablation}
\label{sec:filtering-ablation}
To quantify the contribution of the \textsc{ReMova} filtering pipeline, we run an ablation in which the selected models are re-fine-tuned on unfiltered data drawn from the same source corpora. The ablation run matches the cleaned run in both total volume and per-source proportions, so that any difference in downstream translation quality can be attributed to filtering rather than to changes in training data quantity or composition. All hyperparameters (Table~\ref{tab:hparams}) and the evaluation protocol are kept identical to the main run.

\subsection{Evaluation}
We use the recent BOUQuET evaluation dataset~\cite{andrews-etal-2025-bouquet} for its cross-domain sentence coverage and human-verified translations, which lets us track per-domain performance improvements from fine-tuning. The test split contains 854 unique sentences (198 paragraphs), written in the standard Belarusian orthography, across eight domains: comments, conversation, how-to/instructions, narration, other miscellaneous, reflection, social posts, and web miscellaneous, matching the row labels in Table~\ref{tab:perdomain-beforeafter}. The dev split was used for best-checkpoint selection; all reported metrics come from the held-out test set.

To capture multiple dimensions of translation quality, we report BLEU~\cite{papineni-etal-2002-bleu} and chrF++~\cite{popovic-2017-chrf} as surface-level metrics. ChrF++ is more informative for morphologically rich languages, but reporting both is preferable in low-resource settings, as the two metrics expose different translation artefacts~\cite{kumar2026evaluatingextremelylowresourcemachine}. We additionally report COMET~\cite{rei-etal-2020-comet}, a neural metric that correlates more closely with human judgements. Base LLMs were evaluated with a fixed 3-shot prompt of simple, short demonstration pairs, identical across all base-LLM evaluations.

\section{Results and discussion}
\subsection{Base models}
Base model results are summarised in Table~\ref{tab:results}. Within the dedicated MT family, NLLB-200 outperforms MADLAD-400 by a wide margin at comparable scale (27.2 vs.\ 21.8 BLEU), consistent with NLLB's targeted Belarusian coverage. Among base LLMs, the Gemma family stands apart: Gemma-3 reaches 22.9 BLEU at 4B. The remaining LLMs trail far behind: Llama-3.2 (10.4 BLEU) recovers only lexical fragments, while 
Aya and Qwen-2.5 (3.6--4.6 BLEU) are essentially non-functional despite Aya and Qwen-2.5 being positioned as multilingual models, suggesting scale alone does not compensate for limited Belarusian pretraining exposure. Only NLLB-200 and the Gemma family clear the bar for productive fine-tuning; we advance NLLB-200 to full fine-tuning and Gemma-4-12B-IT together with TranslateGemma-12B-IT to the LoRA track.

\begin{table}[!t]
  \centering
  \small
  \begin{tabular}{@{}lrrrr@{}}
  \toprule
  \textbf{Model} & \textbf{Size} & \textbf{BLEU} & \textbf{chrF++} & \textbf{COMET} \\
  \midrule
  \multicolumn{5}{@{}l@{}}{\textit{Dedicated MT models}} \\[0.3em]
  NLLB-200              & 3B          & \textbf{27.2} & \textbf{51.4} & \textbf{86.7} \\
  MADLAD-400            & 3B            & 21.8 & 41.6 & 81.2 \\
  \midrule
  \multicolumn{5}{@{}l@{}}{\textit{Base LLMs}} \\[0.3em]
  Gemma-4$^{\dagger}$          & 12B    & \textbf{25.2} & 49.1 & 84.8 \\
  TranslateGemma$^{\dagger\ddagger}$ & 12B & 25.0 & \textbf{51.7} & \textbf{89.6} \\
  Gemma-3               & 4B            & 22.9 & 46.4 & 84.5 \\
  Llama-3.2             & 3B            & 10.4 & 30.7 & 69.6 \\
  Aya                   & 3B         &  4.6 & 19.8 & 50.5 \\
  Qwen-2.5              & 3B            &  3.6 & 18.5 & 48.0 \\
  \bottomrule
  \end{tabular}
\caption{English$\rightarrow$Belarusian on BOUQuET test; best per family in bold. Base LLMs use a fixed 3-shot prompt. $^{\dagger}$Above the 4B screening cap. $^{\ddagger}$Native chat template (not 3-shot).}
  \label{tab:results}
  \end{table}
\subsection{Fine-tuned models}
\label{sec:results-ft}
Fine-tuning results for NLLB, Gemma-4, and TranslateGemma are presented in Table~\ref{tab:perdomain-beforeafter}, alongside a matched ablation in which the same models are re-fine-tuned on unfiltered data from the same source corpora (\S\ref{sec:filtering-ablation}, Table~\ref{tab:hparams}).


\begin{table*}[!tb]
\centering
\small
\setlength{\tabcolsep}{3pt}
\renewcommand{\arraystretch}{0.92}
\begin{tabular}{l@{\hspace{2em}}r@{\hspace{2em}}ccc@{\hspace{2em}}ccc@{\hspace{2em}}ccc}
\toprule
 & & \multicolumn{3}{c}{\hspace{-2em}\textbf{BLEU}} & \multicolumn{3}{c}{\hspace{-1.75em}\textbf{chrF++}} & \multicolumn{3}{c}{\textbf{COMET}} \\[0.3em]
Domain & $n$ & base & unfilt. & +FT & base & unfilt. & +FT & base & unfilt. & +FT \\
\midrule
\multicolumn{11}{l}{\textit{NLLB-200-3.3B}} \\[0.3em]
Comments      & 73  & 25.22 & 31.18 & \textbf{32.83} & 48.17 & 51.23 & \textbf{52.72} & 83.07 & 85.76 & \textbf{86.40} \\
Conversation  & 186 & 26.83 & 34.92 & \textbf{35.18} & 49.43 & 53.62 & \textbf{54.66} & 85.87 & 89.27 & \textbf{90.07} \\
How-to/instr. & 107 & 25.77 & 24.29 & \textbf{27.68} & 50.65 & 50.42 & \textbf{53.19} & 84.10 & 83.66 & \textbf{85.33} \\
Narration     & 98  & 27.41 & 30.95 & \textbf{32.57} & 51.68 & \textbf{56.08} & 56.04 & 85.33 & 87.29 & \textbf{88.56} \\
Other misc.   & 95  & 28.98 & 30.03 & \textbf{32.38} & 53.46 & 54.70 & \textbf{55.46} & 88.99 & 89.79 & \textbf{91.13} \\
Reflection    & 88  & 25.95 & 24.42 & \textbf{27.43} & 52.39 & 51.86 & \textbf{52.91} & 89.09 & 88.78 & \textbf{89.56} \\
Social posts  & 104 & 30.35 & 30.58 & \textbf{32.30} & 52.87 & 53.26 & \textbf{53.93} & 88.39 & 88.26 & \textbf{89.88} \\
Web misc.     & 103 & 24.20 & 24.67 & \textbf{26.30} & 51.17 & 52.82 & \textbf{53.86} & 88.65 & 89.08 & \textbf{90.77} \\
\addlinespace
\textit{Overall} & 854 & 27.24 & 29.23 & \textbf{31.22} & 51.35 & 53.09 & \textbf{54.19} & 86.67 & 87.90 & \textbf{89.12} \\
\midrule
\multicolumn{11}{l}{\textit{Gemma-4-12B-IT}} \\[0.3em]
Comments      & 73  & 26.49 & 28.47 & \textbf{28.66} & 48.06 & \textbf{49.87} & 49.77 & 83.58 & 82.33 & \textbf{86.65} \\
Conversation  & 186 & 32.54 & 31.61 & \textbf{34.74} & 50.64 & 52.12 & \textbf{52.67} & 86.63 & 87.48 & \textbf{87.78} \\
How-to/instr. & 107 & 21.60 & 23.81 & \textbf{28.88} & 48.56 & 51.13 & \textbf{52.88} & 82.89 & 85.04 & \textbf{86.37} \\
Narration     & 98  & 24.41 & 27.97 & \textbf{33.46} & 49.31 & 53.55 & \textbf{54.92} & 82.25 & 87.09 & \textbf{88.80} \\
Other misc.   & 95  & 24.64 & 28.29 & \textbf{30.17} & 48.75 & 53.07 & \textbf{53.95} & 85.35 & 89.19 & \textbf{90.41} \\
Reflection    & 88  & 22.20 & 23.45 & \textbf{29.73} & 48.24 & 50.65 & \textbf{53.03} & 84.49 & 86.98 & \textbf{90.29} \\
Social posts  & 104 & 26.09 & 30.75 & \textbf{32.64} & 50.37 & 53.57 & \textbf{55.69} & 86.22 & 87.59 & \textbf{90.17} \\
Web misc.     & 103 & 19.22 & 23.83 & \textbf{28.05} & 47.53 & 51.80 & \textbf{54.56} & 85.06 & 88.73 & \textbf{90.93} \\
\addlinespace
\textit{Overall} & 854 & 25.23 & 27.78 & \textbf{31.40} & 49.08 & 52.11 & \textbf{53.64} & 84.80 & 86.99 & \textbf{88.85} \\
\midrule
\multicolumn{11}{l}{\textit{TranslateGemma-12B-IT}} \\[0.3em]
Comments      & 73  & 25.33 & 26.77 & \textbf{30.93} & 50.65 & 49.50 & \textbf{53.08} & \textbf{88.84} & 83.16 & 87.57 \\
Conversation  & 186 & 28.78 & 32.82 & \textbf{37.90} & 50.18 & 51.60 & \textbf{55.19} & 88.83 & 87.90 & \textbf{89.01} \\
How-to/instr. & 107 & 24.47 & 25.66 & \textbf{32.65} & 52.69 & 52.30 & \textbf{57.45} & 87.90 & 84.49 & \textbf{88.57} \\
Narration     & 98  & 24.54 & 29.42 & \textbf{34.09} & 52.09 & 54.96 & \textbf{57.67} & 88.73 & 87.41 & \textbf{89.64} \\
Other misc.   & 95  & 27.79 & \textbf{31.49} & 31.40 & 54.05 & \textbf{56.41} & 53.88 & 90.95 & 89.33 & \textbf{91.79} \\
Reflection    & 88  & 20.29 & 24.29 & \textbf{27.80} & 50.33 & 51.34 & \textbf{53.38} & 90.52 & 88.29 & \textbf{91.17} \\
Social posts  & 104 & 24.98 & 28.52 & \textbf{32.15} & 51.96 & 52.60 & \textbf{55.34} & 90.69 & 86.84 & \textbf{90.75} \\
Web misc.     & 103 & 21.49 & 23.62 & \textbf{28.52} & 51.59 & 52.14 & \textbf{55.67} & 91.24 & 89.92 & \textbf{92.21} \\
\addlinespace
\textit{Overall} & 854 & 24.98 & 28.25 & \textbf{32.63} & 51.73 & 52.73 & \textbf{55.36} & 89.63 & 87.33 & \textbf{90.03} \\
\bottomrule 
\end{tabular}
\caption{Per-domain English$\rightarrow$Belarusian test scores. Columns: \emph{base} (pre-fine-tune baseline), \emph{unfilt.} (fine-tuned on unfiltered data from the same sources), \emph{+FT} (fine-tuned on the \textsc{ReMova}-cleaned corpus). Best per row in bold.}
\label{tab:perdomain-beforeafter}
\end{table*}

All three fine-tuned models cluster within a narrow band on aggregate metrics, with TranslateGemma leading both BLEU (32.63) and chrF++ (55.36), Gemma-4 close behind on BLEU (31.40), and NLLB slightly ahead of Gemma-4 on chrF++ (Table~\ref{tab:perdomain-beforeafter}). The close finish is itself notable: a general-purpose LLM (Gemma-4-12B) with LoRA fine-tuning on cleaned data reaches parity with a fully fine-tuned dedicated MT model (NLLB-3.3B) without any translation-specific pretraining objective.

For TranslateGemma specifically, fine-tuning yields a large lexical improvement ($+7.65$ BLEU, $+3.63$ chrF++). Per-domain scores, presented in Table~\ref{tab:perdomain-beforeafter}, show that
fine-tuning improves BLEU in every domain, but by markedly uneven margins: gains range from $+9.55$ BLEU on Narration and $+9.12$ on Conversation down to $+3.61$ on Other misc., the only domain where chrF++ does not improve ($-0.17$).

Per-domain BLEU gains from fine-tuning vary across models, though Narration improves substantially for all three ($+5.2$ for NLLB, $+9.1$ for Gemma-4, $+9.6$ for TranslateGemma). This is plausibly driven by OpenSubtitles, which contributes 35\% of the training data and was upsampled 2$\times$ to enrich the spoken register (\S\ref{sec:methodology}). The distribution of smaller gains differs by model: NLLB benefits least on Reflection and How-to; Gemma-4 on Comments and Conversation and TranslateGemma on Other miscellaneous. Diversifying the stylistic range of training data is a natural direction for future work.

\subsection{Filtering ablation}
To measure the contribution of the \textsc{ReMova} cleaning pipeline in isolation, we compare each model's cleaned fine-tune against a matched fine-tune on unfiltered data from the same source corpora (Table~\ref{tab:perdomain-beforeafter}, ``unfilt.''\ column). Cleaning delivers $+1.99$ BLEU / $+1.10$ chrF++ for NLLB, $+3.62$ / $+1.53$ for Gemma-4, and $+4.38$ / $+2.63$ for TranslateGemma (COMET deltas follow the same ordering). The pattern appears to split along the encoder-decoder-MT versus decoder-only-LLM divide rather than along model size or degree of translation specialisation: NLLB, whose pretraining objective is already close to supervised parallel translation, gains roughly two times less from cleaning than either decoder-only LLM. Notably, TranslateGemma's translation-specialisation does not close this  gap; despite reaching the highest overall quality after fine-tuning, it shows the largest cleaning gain of the three.

At the per-domain level, some cases sharpen this story. For Gemma-4 Conversation, unfiltered fine-tuning actually drops BLEU below the 3-shot base (31.61 vs.\ 32.54), while cleaned fine-tuning delivers a clear gain (34.74): fine-tuning on noisy data can be net-negative for that domain. On BLEU, cleaning wins every domain for every model except TranslateGemma@Other miscellaneous (−0.09); on chrF++, cleaning wins every domain except three isolated cells (NLLB@Narration, Gemma-4@Comments, and TranslateGemma@Other miscellaneous), consistent with a slightly wider surface-form vocabulary in the unfiltered pool, helping character n-gram matching on domains dominated by rare terms.

As shown in our pipeline results (Table~\ref{tab:clean-tatoeba}), even datasets considered ``cleaned'' upstream retain substantial noise and require further downstream filtering; we leave deeper investigation to future work. Taken together, the ablation supports the paper's data-quality thesis: for the English--Belarusian pair, systematic filtering of parallel data is a first-order lever on translation quality rather than a secondary tweak.

\section{Related Work}

\paragraph{Multilingual translation.}
While high-resource languages benefit from large parallel corpora suitable for training and evaluating translation models, many low-resource languages long lacked sufficient support. Yet, openly released encoder-decoder systems have been progressively expanding the language coverage: OPUS-MT provided per-pair models trained on
the OPUS collection \citep{tiedemann-2012-parallel, tiedemann-thottingal-2020-opus};
M2M-100 introduced many-to-many translation across 100
languages \citep{3546258.3546365}; NLLB scaled this recipe to more than 200
languages \citep{nllbteam2022languageleftbehindscaling}; MADLAD-400 extended
coverage past 400 \citep{kudugunta2023madlad400multilingualdocumentlevellarge};
and the recently-announced Omnilingual MT targets
1\,600 languages \citep{omnilingualmtteam2026omnilingualmtmachinetranslation}.

\paragraph{LLM-based translation.}
In parallel with the development of dedicated encoder-decoder systems, the scaling of LLMs has opened an alternative path to translation. Off-the-shelf language models were found to already translate well between high-resource languages using few-shot learning \citep{3618408.3618846}, but fall behind dedicated systems when translating low-resource languages \citep{robinson-etal-2023-chatgpt}. However, fine-tuning the LLMs can close this gap -- with approaches like the Advanced Language Model-based translator \citep[ALMA;][]{xu2024paradigmshiftmachinetranslation}. Finetuned from Llama 2 \citep{touvron2023llama2openfoundation}, ALMA outperformed the 54B-parameter NLLB model despite its much smaller size of 7B parameters. Its training follows a two-stage recipe: continued pretraining on monolingual data in the target languages, followed by fine-tuning on a modest amount of high-quality bilingual corpus. The authors argue that excessive training on parallel data washes out the LLM's pre-existing knowledge, although a subsequent study \citep{iyer-etal-2024-quality} shows that this finding does not hold for low-resource languages with limited pretraining coverage. Subsequent work has further refined this recipe: Tower mixes translation with related tasks during fine-tuning, further improving translation quality \citep{alves2024toweropenmultilinguallarge, rei-etal-2024-tower}. Its successor, Tower+, scales data quality, quantity, and model size and additionally employs reinforcement learning to refine the generated outputs \citep{rei-etal-2026-tower}. TranslateGemma applies a similar recipe to the open Gemma 3 models, providing one of our fine-tuning bases \citep{finkelstein2026translategemmatechnicalreport}.

\paragraph{WMT General MT Task.}
WMT has been focusing on machine translation of low-resource languages for many years: in 2022, the shared task included Livonian and Yakut languages \citep{kocmi-etal-2022-findings}, and the range of low-resource languages has further expanded in the following iterations. In 2023, translation systems based on LLMs started to be competitive in WMT \citep{kocmi-etal-2023-findings}, and then in 2024, finetuned LLM systems started outperforming the standard encoder-decoder approaches \citep{kocmi-etal-2024-findings}. This trend aligns with our findings that the best performance on English-Belarusian translation is achieved with a finetuned LLM.

\paragraph{Data quality and filtering.}
Much of the parallel data behind multilingual MT is web-mined with cross-lingual sentence embeddings: LASER-based mining \citep{artetxe-schwenk-2019-margin} produced corpora such as CCAligned, CCMatrix or WikiMatrix \citep{el-kishky-etal-2020-ccaligned,schwenk-etal-2021-ccmatrix,schwenk-etal-2021-wikimatrix}, and NLLB mined its low-resource bitext with distilled LASER~3 encoders \citep{nllbteam2022languageleftbehindscaling}. We are using a later improvement of LASER called BLASER in this study \citep{seamlessm4t2023blaser}. These heuristically-aligned corpora are known for systematic quality issues, especially for low-resource languages \citep{kreutzer-etal-2022-quality}. Proper filtering of these corpora has thus become an essential step in training NMT systems, supported by tools such as Zipporah \citep{xu-koehn-2017-zipporah}, Bicleaner and Bifixer, its neural successor \citep{ramirez-sanchez-etal-2020-bifixer, zaragoza-bernabeu-etal-2022-bicleaner}, as well as the modular OpusFilter and OpusCleaner toolkits~\cite{aulamo-etal-2020-opusfilter, bogoychev2023opuscleaneropustraineropensource}. These pipelines combine many approaches but remain largely language-agnostic; our work follows the same overall design while adding Belarusian-specific orthographic normalisation and spellchecking that generic tools cannot provide.


\section{Conclusion}
In this work, we presented \textsc{ReMova}, a Belarusian-specific data-cleaning pipeline, and fine-tuned translation models for the English--Belarusian language pair across both the encoder-decoder and decoder-only paradigms. The pipeline combines language identification, Unicode normalisation, spelling correction with Wikificator and LanguageTool, detection of misaligned parallel segments, and filtering or replacement of \textit{Taraškievica} and obvious non-Belarusian forms; applying it exposed substantial noise even in datasets we initially considered higher-quality. On the cleaned corpus, full fine-tuning of NLLB-200-3.3B, LoRA fine-tuning of Gemma-4-12B-IT, and LoRA fine-tuning of TranslateGemma-12B-IT produced comparable performance on the BOUQuET test set, with TranslateGemma leading on every reported metric (32.63 BLEU, 55.36 chrF++, 90.03 COMET) and Gemma-4 reaching parity with the fully fine-tuned dedicated MT baseline without any translation-specific pretraining objective. A matched ablation in which each model was re-fine-tuned on unfiltered data from the same sources showed substantial BLEU gains from filtering for all three models, with the decoder-only LLMs (Gemma-4 $+3.62$, TranslateGemma $+4.38 $) gaining roughly two times more than the dedicated encoder-decoder MT system NLLB ($+1.99$); chrF++ and COMET follow the same architectural split. Taken together, these results support the hypothesis that for the English--Belarusian pair, data quality is a first-order lever on translation quality.

Several directions remain for future work. Scaling back-translation beyond the current 2{,}300 curated HPLT segments (using a larger monolingual pool, additional back-translation models, and pivoting through related higher-resource pairs such as Belarusian--Russian) could broaden training coverage beyond OPUS-derived sources. Diversifying the stylistic range of training data would likely close the per-domain gap on formal and reflective content. Finally, human evaluation and statistical-significance testing would give the model rankings and ablation deltas reported here a firmer footing than metric-based comparison alone.

\section*{Limitations}
We did not conduct a thorough human evaluation, which would provide the strongest signal of translation quality and help analyse mistakes. Metrics such as BLEU underestimate quality for morphologically rich languages like Belarusian and may correlate weakly with human judgements. COMET was trained on human judgements dominated by high-resource language pairs and has not been meta-evaluated on Belarusian; absolute COMET values reported here should be treated as internally consistent within this paper rather than directly comparable to published COMET numbers on other language pairs. We did not test the statistical significance of our results, which would be especially valuable for the smaller ablation gaps and the tight ordering between models.

Some larger or potentially more capable models for Belarusian translation were not evaluated. We did not benchmark against commercial systems such as Yandex Translate or Google Translate, which cover Belarusian; comparison with a strong commercial baseline would help calibrate our absolute scores against production-quality MT and is left for future work. The few-shot examples used for base-LLM evaluation may have affected performance and should be varied in future evaluations to test prompt sensitivity.

Compute constraints motivated the LoRA-only setup for the LLM fine-tuning and the 4B parameter cap on the initial base-LLM screen; both choices may have affected the absolute scores and the cross-model comparisons. The hyperparameter search was limited, and we cannot guarantee the chosen settings were optimal.

\section*{Use of AI assistants}
During the preparation of this work, the authors used Claude (Anthropic, \texttt{claude-opus-4-7, claude-opus-4-8, claude-opus-5-0}) and GPT (OpenAI, \texttt{gpt-5.6-sol}) to assist with writing the paper, specifically for refining wording and proof-reading. All AI-assisted output was reviewed and edited by the authors, who take full responsibility for the correctness, design decisions, and content of the submitted work. No sensitive or personal data was entered into the tools, and all data handling complied with the University of Oslo's data regulations.

\FloatBarrier

\bibliography{custom}

\newpage

\appendix

\section{Belarusian tokenisation across tokenisers}
\label{app:tokeniser}

To get insights on Belarusian cyrillic text tokenisation, we constructed the evaluation sample of 1~013~384 characters from the Belarusian Wikipedia articles listed by \textit{Special:LongPages}, processing them in descending page-length order, and made a benchmarking on several tokenisers. Articles were retrieved as plain text through the MediaWiki API. We retained only paragraphs classified as Belarusian whose alphabetic characters belonged exclusively to the Belarusian alphabet, and concatenated complete article extracts until the sample exceeded one million characters. The resulting text was passed unchanged to each tokeniser.

For the Benchmark we used several tokenisers, available at The Tokeniser Playground~\cite{xenova_tokenizer_playground} and the tokeniser from CLARIN project, specifically made for Belarusian language~\cite{hetsevich2022_kblp}.

 \begin{table}[!th]
  \centering
  \small
  \begin{tabular}{@{}lrr@{}@{}}
  \toprule
  \textbf{Tokeniser} & \textbf{Token} & \textbf{Token/} \\
  & \textbf{count} & \textbf{character} \\
  \midrule
    \texttt{Xenova/gpt-4} & 663445 & 0,655 \\
    \texttt{Xenova/claude-tokeniser} & 684587 & 0,676 \\
    \texttt{Xenova/gemma-tokeniser} & 448819 & 0,443 \\
    \texttt{Xenova/llama-tokeniser} & 519320 & 0,512 \\
    CLARIN Tokeniser & \textbf{326903} & \textbf{0,323} \\
  \bottomrule
  \end{tabular}
  \caption{Token counts for the same Belarusian text sample of 1~013~384 characters and 263~707 words from Belarusian wikipedia \textit{Special:LongPages} at the moment of 2~August~2026.}
  \label{tab:tokeniser-bel}
  \end{table}

\end{document}